\documentclass{article}

\usepackage{arxiv}
\usepackage{amsmath}
\usepackage{amssymb}
\usepackage{cite}
\usepackage{colortbl}
\usepackage{array}
\usepackage[hidelinks]{hyperref}
\usepackage{draftwatermark}
\SetWatermarkText{PREPRINT}
\SetWatermarkLightness{0.95}
\graphicspath{{figs/}}

\begin{document}

\title{3W Dataset 2.0.0: a realistic and public dataset with rare undesirable real events in oil wells}

\author{%
    \begin{minipage}{0.90\textwidth}
        \centering
        \href{https://orcid.org/0000-0001-6243-4590}{\includegraphics[scale=0.06]{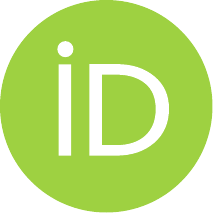}\hspace{1mm}Ricardo Emanuel Vaz Vargas} \thanks{Corresponding author (\texttt{ricardo.vargas@petrobras.com.br}, \href{https://www.linkedin.com/in/ricardovvargas}{https://www.linkedin.com/in/ricardovvargas}).} $^{,1}$, \href{https://orcid.org/0000-0002-7279-3981}{\includegraphics[scale=0.06]{orcid.pdf}\hspace{1mm}Afrânio José de Melo Junior} $^1$, \href{https://orcid.org/0000-0002-2297-7395}{\includegraphics[scale=0.06]{orcid.pdf}\hspace{1mm}Celso José Munaro} $^3$, \href{https://orcid.org/0000-0002-3833-321X}{\includegraphics[scale=0.06]{orcid.pdf}\hspace{1mm}Cláudio Benevenuto de Campos Lima} $^1$, \href{https://orcid.org/0000-0003-3894-8190}{\includegraphics[scale=0.06]{orcid.pdf}\hspace{1mm}Eduardo Toledo de Lima Junior} $^6$, \href{https://orcid.org/0009-0004-7337-2013}{\includegraphics[scale=0.06]{orcid.pdf}\hspace{1mm}Felipe Muntzberg Barrocas} $^5$, \href{https://orcid.org/0000-0002-5444-1974}{\includegraphics[scale=0.06]{orcid.pdf}\hspace{1mm}Flávio Miguel Varejão} $^4$, \href{https://orcid.org/0000-0002-5238-4910}{\includegraphics[scale=0.06]{orcid.pdf}\hspace{1mm}Guilherme Fidelis Peixer} $^7$, \href{https://orcid.org/0009-0004-0132-3124}{\includegraphics[scale=0.06]{orcid.pdf}\hspace{1mm}Igor de Melo Nery Oliveira} $^6$, \href{https://orcid.org/0000-0002-8753-6670}{\includegraphics[scale=0.06]{orcid.pdf}\hspace{1mm}Jader Riso Barbosa Jr.} $^2$, \href{https://orcid.org/0000-0003-2682-9952}{\includegraphics[scale=0.06]{orcid.pdf}\hspace{1mm}Jaime Andrés Lozano Cadena} $^7$, \href{https://orcid.org/0009-0001-7954-8540}{\includegraphics[scale=0.06]{orcid.pdf}\hspace{1mm}Jean Carlos Dias de Araújo} $^1$, \href{https://orcid.org/0000-0001-7596-622X}{\includegraphics[scale=0.06]{orcid.pdf}\hspace{1mm}João Neuenschwander Escosteguy Carneiro} $^5$, \href{https://orcid.org/0000-0001-7761-7293}{\includegraphics[scale=0.06]{orcid.pdf}\hspace{1mm}Lucas Gouveia Omena Lopes} $^6$, \href{https://orcid.org/0000-0002-4975-2858}{\includegraphics[scale=0.06]{orcid.pdf}\hspace{1mm}Lucas Pereira de Gouveia} $^6$, \href{https://orcid.org/0000-0002-6416-5061}{\includegraphics[scale=0.06]{orcid.pdf}\hspace{1mm}Mateus de Araujo Fernandes} $^1$, \href{https://orcid.org/0000-0003-4176-8158}{\includegraphics[scale=0.06]{orcid.pdf}\hspace{1mm}Matheus Lima Scramignon} $^5$, \href{https://orcid.org/0000-0003-3177-4028}{\includegraphics[scale=0.06]{orcid.pdf}\hspace{1mm}Patrick Marques Ciarelli} $^3$, \href{https://orcid.org/0000-0003-1870-4156}{\includegraphics[scale=0.06]{orcid.pdf}\hspace{1mm}Rodrigo Castello Branco} $^5$, \href{https://orcid.org/0000-0002-3418-7606}{\includegraphics[scale=0.06]{orcid.pdf}\hspace{1mm}Rogério Leite Alves Pinto} $^1$
    \end{minipage}
\\
\\
    \begin{minipage}{0.85\textwidth}
        \centering
        \small
        $^1$ Petróleo Brasileiro S.A. (Petrobras), Rio de Janeiro, Brazil\\
        $^2$ Craft \& Hawkins Department of Petroleum Engineering, Louisiana State University, Baton Rouge, United States of America\\
        $^3$ Department of Electrical Engineering, Federal University of Espírito Santo (UFES), Vitória, Brazil\\
        $^4$ Department of Informatics, Federal University of Espírito Santo (UFES), Vitória, Brazil\\
        $^5$ HybridAI, Rio de Janeiro, Brazil\\
        $^6$ Laboratório de Computação Científica e Visualização (LCCV), Federal University of Alagoas (UFAL), Maceió, Brazil\\
        $^7$ POLO - Research Laboratories for Emerging Technologies in Cooling and Thermophysics, Department of Mechanical Engineering, Federal University of Santa Catarina (UFSC), Florianópolis, Brazil\\
    \end{minipage}
}

\maketitle

\begin{abstract}
In the oil industry, undesirable events in oil wells can cause economic losses, environmental accidents, and human casualties. Solutions based on Artificial Intelligence and Machine Learning for Early Detection of such events have proven valuable for diverse applications across industries. In 2019, recognizing the importance and the lack of public datasets related to undesirable events in oil wells, Petrobras developed and publicly released the first version of the 3W Dataset, which is essentially a set of Multivariate Time Series labeled by experts. Since then, the 3W Dataset has been developed collaboratively and has become a foundational reference for numerous works in the field. This data article describes the current publicly available version of the 3W Dataset, which contains structural modifications and additional labeled data. The detailed description provided encourages and supports the 3W community and new 3W users to improve previous published results and to develop new robust methodologies, digital products and services capable of detecting undesirable events in oil wells with enough anticipation to enable corrective or mitigating actions.
\end{abstract}

\section{Background \& Summary}

Undesirable events cause different types of damage to the oil industry, including economic losses, environmental accidents, and human casualties \cite{ref01}.

Abnormal Event Management (AEM) refers to systematic detection, diagnosis, and mitigation of unexpected or irregular events within complex industrial systems \cite{ref02}. In the oil industry, where operational safety, environmental protection, and economic performance are tightly coupled, AEM is crucial for minimizing the impact of undesirable events. The integration of Artificial Intelligence (AI) and Machine Learning (ML) based solutions into AEM have shown promise for Early Detection \cite{ref03} of undesirable events across different industries \cite{ref04}. By analyzing large streams of operational data, such as pressure, temperature, vibration, and flow rates, algorithms can identify subtle patterns that precede abnormal conditions, allowing early intervention and preventive maintenance strategies \cite{ref05}. An essential requirement for the success of this type of approach is the use of high-quality datasets \cite{ref06}\cite{ref07}.

This need was recognized by Petrobras — the largest oil company in Brazil, operating in the exploration, production, refining, marketing, and transportation of oil, natural gas, and energy \cite{ref08} — which developed the 3W Dataset and published its first version in 2019, as described in detail by Vargas et al. \cite{ref09}. 

The 3W Dataset is a collection of Multivariate Time Series (MTS) \cite{ref10}\cite{ref11}, referred to as instances, which have been labeled by experts from Petrobras and its partners. The name \textbf{\textit{3W}} was chosen because this dataset comprises instances derived from 3 different sources (real, simulated, and hand-drawn) which contains undesirable events that occur in oil \textbf{\textit{W}}ells. Each instance may either represent 100\% of the data associated with normal operating conditions or contain data partially related to a specific type of undesirable event. The core idea behind this dataset is to enable the learning (modeling) of temporal patterns or signatures across multiple variables, distinguishing between normal conditions and various types of undesirable events.

The main features of the 3W Dataset are as follows. Its real instances correspond to data collected directly from real industrial environments. Characteristics such as frozen variables, missing variables, and outliers are intentionally left untreated. This approach aims to encourage and enable the development of methodologies and digital products capable of dealing with real-world challenges. By preserving these typical characteristics, a high-quality and useful dataset is generated \cite{ref12}. Simulated instances have been added because some types of undesirable events are rare in real-world operations. Hand-drawn instances have been added to address events that are not only rare but also challenging to simulate. Experts can use their understanding of variable behaviors during such events to manually create these instances to mirror real-world conditions as closely as possible.

Due to its features, the 3W Dataset can also serve as a foundational resource for training basic models in Transfer Learning frameworks \cite{ref13}. These frameworks rely on a large and generalized dataset as a reference or starting point for training models to address different but related challenges, particularly when an adequate training dataset for the target task is unavailable. This approach is widely used in Deep Learning \cite{ref14}, especially in scenarios requiring large amounts of training data.

The 3W Dataset is managed using Semantic Versioning \cite{ref15}, with its initial release designated as version 1.0.0.

Since its initial release, the 3W Dataset has been explored by several people who make up the 3W Community \cite{ref16}, including independent professionals and representatives of research institutions, startups, companies, and oil operators from different countries. 

The 3W Community has contributed to the development and publication of numerous works, forming a substantial scientific framework focused on the Early Detection of undesirable events in oil wells. This framework is composed of books, conference papers, doctoral theses, final graduation projects, journal articles, master's degree dissertations, repository articles, and specialization monographs. Publications identified so far that cite the 3W Dataset are listed in the 3W Project repository \cite{ref17}.

In 2022, Petrobras officially launched the 3W Project as the inaugural pilot of the Open Lab Module within the Connections for Innovation Program \cite{ref18}. The purpose of this module is to encourage open and collaborative project development on the Internet, particularly through the GitHub platform \cite{ref19}. Since then, the 3W Dataset has been maintained and developed in its dedicated corporate Git repository \cite{ref17} on GitHub.

As part of the 3W Project, the same Git repository also hosts the 3W Toolkit, a software package developed in Python 3 \cite{ref20}. The purpose of this toolkit is to facilitate and encourage the exploration of the 3W Dataset, as well as the development and comparison of different methodological approaches.

In addition to its two main resources — the 3W Dataset and the 3W Toolkit — the 3W Project repository also provides:

\begin{itemize}
    \item A detailed description of motivation, strategy, ambition, governance, and other aspects of the 3W Project;
    \item A list of at least 100 published works that cite the 3W Dataset;
    \item Specifications of priority challenges (benchmarks);
    \item The 3W Project contributing guide;
    \item Information about the 3W Community;
    \item Overviews of the 3W Dataset developed by the 3W Community;
    \item The 3W Community code of conduct;
    \item Release notes for the published versions of the 3W Dataset.
\end{itemize}

Since 2022, the 3W Dataset has been evolved under Petrobras' leadership, and its current publicly available version designated as 2.0.0.

This data article describes the 3W Dataset 2.0.0 and summarizes the advancements incorporated since version 1.0.0. The detailed description is intended to encourage and support the established 3W Community and also new users to improve previously published results and develop innovative methodologies, digital products and services. These advancements aim to enable Early Detection of undesirable events in oil wells, allowing for timely corrective or mitigating actions.

\section{Methods}

In summary, the 3W Dataset 2.0.0 is composed of three types of instances generated by three methods, one per type of instance. These methods are described in this section and are based on the mathematical definition of MTS presented in the following subsection. According to this definition and to support both the 3W Project and this article, a nomenclature was developed and is detailed in the subsequent subsection. The types of instances and the methods associated with them are described in their own subsections. Some characteristics are common to the three methods and are, therefore, detailed in a separate subsection.

\subsection{Mathematical Definition of Multivariate Time Series}

The chosen definition for MTS is the same as that used in the article that published the 3W Dataset 1.0.0 \cite{ref09}. This definition is reproduced below.

A dataset \textit{DS} is a set of $m$ MTS ($S^{i} | i = \{1, 2, \dots, m\}$, $\forall m \in \mathbb{Z}$, and $m > 1$) and is defined as \textit{DS} = $\{\textit{S}^1, \textit{S}^2, \dots, \textit{S}^m\}$. Each MTS $i$ is an instance composed of a set of $n$ univariate time series ($x^{i}_{j} | j = \{1, 2, \dots, n\}$, $\forall n \in \mathbb{Z}$, and $n > 1$) (also referenced as process variable or just variable), and is defined as \textit{S}$^{i}$ = $\{x^{i}_{1}, x^{i}_{2}, \dots, x^{i}_{n}\}$. Each variable $j$ that composes an MTS $i$ is an ordered temporal sequence of $p_{i}$ observations taken at the time $t$ ($x^{i}_{j,t} | t = \{1, 2, \dots, p_{i}\}$, $\forall p_{i} \in \mathbb{Z}$, and $p_{i} > 1$). Therefore, each MTS $i$ is considered in this work as a matrix defined as $S^{i}$ = $\{x^{i}_{1,1}, x^{i}_{2,1}, \dots, x^{i}_{n,1}; x^{i}_{1,2}, x^{i}_{2,2}, \dots, x^{i}_{n,2}; x^{i}_{1,p_{i}}, x^{i}_{2,p_{i}}, \dots, x^{i}_{n,p_{i}}\}$.

Note that all instances have a fixed number of variables $n$ all instances are composed of the same $n$ variables; each instance can be composed of any number of observations $p_{i}$; all $n$ variables of an instance $i$ have a fixed number of observations $p_{i}$; and different instances can be composed of different numbers of observations.

\subsection{Nomenclature}

The nomenclature used in this work is derived from the definition of MTS presented in the previous subsection. The specific terms and their descriptions are presented in Table \ref{tab:table1}.

\begin{table}[hbt!]
\centering
\caption{Terms that make up the nomenclature used in this work.}
{\renewcommand{\arraystretch}{1.5}
\begin{tabular}{ | l | l | } 
\hline
\rowcolor[gray]{0.8} 
\multicolumn{1}{| c |}{\textbf{Term}} & 
\multicolumn{1}{ c |}{\textbf{Meaning}} \\
\hline
Variable & 
\begin{minipage}[t]{0.8\columnwidth}%
A physical quantity at a specific point in the production system of a particular well, from which measurements are acquired to produce a univariate time series $j$: $x_{j}$ 
\vspace{2 mm}
\end{minipage}
\\
\hline
Timestamp & 
\begin{minipage}[t]{0.8\columnwidth}%
Instant $t$ (date + time) at which values are acquired or generated and then associated with variables: YYYY-MM-DD HH:MM:SS
\vspace{2 mm}
\end{minipage}
\\
\hline
Observation & 
\begin{minipage}[t]{0.8\columnwidth}%
Vector with values from $n$ variables of a single instance $i$ acquired at a timestamp $t$: $\{x^{i}_{1,t}, x^{i}_{2,t}, \dots, x^{i}_{n,t}\}$ 
\vspace{2 mm}
\end{minipage}
\\
\hline
Label &
\begin{minipage}[t]{0.8\columnwidth}%
 Annotation provided by an expert regarding the well condition in terms of a particular property. The 3W Dataset 2.0.0 includes two types of labels which are defined below: class label and state label. The labeling process is explained in Common Characteristics Among the Methods Subsection 
\vspace{2 mm}
\end{minipage}
\\
\hline
Class label & 
\begin{minipage}[t]{0.8\columnwidth}%
Annotation provided by an expert regarding the well condition in terms of normality or the occurrence of an undesirable event. See additional explanation in Common Characteristics Among the Methods Subsection 
\vspace{2 mm}
\end{minipage}
\\
\hline
State label & 
\begin{minipage}[t]{0.8\columnwidth}%
Annotation provided by an expert regarding the well condition in terms of operational status. See additional explanation in Common Characteristics Among the Methods Subsection 
\vspace{2 mm}
\end{minipage}
\\
\hline
Sample & 
\begin{minipage}[t]{0.8\columnwidth}%
Part of an MTS, including all $n$ variables and all observations between two timestamps 
\vspace{2 mm}
\end{minipage}
\\
\hline
Period & 
\begin{minipage}[t]{0.8\columnwidth}%
{\renewcommand{\arraystretch}{1.5}
A sample that respects the following two conditions: all its observations are labeled with the same class label, and it is not contained in another temporally larger sample with all observations labeled with the same class label. In other words, a period is the largest possible sample whose observations are labeled with the same class label}
\vspace{2 mm}
\end{minipage}
\\
\hline
Instance & 
\begin{minipage}[t]{0.8\columnwidth}%
Collection of temporally sequential periods associated with a specific well 
\vspace{2 mm}
\end{minipage}
\\
\hline
Type of event & 
\begin{minipage}[t]{0.8\columnwidth}%
Operational state in which a well can be found, including normality, failures, and undesired states
\vspace{2 mm}
\end{minipage}
\\
\hline
Dataset & 
\begin{minipage}[t]{0.8\columnwidth}%
Set of instances with multiple types of events 
\vspace{2 mm}
\end{minipage}
\\
\hline
\end{tabular}}
\label{tab:table1}
\end{table}

\subsection{Types of Instances}

As mentioned at the beginning of this section, the 3W Dataset 2.0.0 is composed of three types of instances, denoted as real, simulated, and hand-drawn.

Each type of instance has been defined based on the origin of its data. Real instances were sourced from different Petrobras' Plant Information Management Systems (PIMS) \cite{ref21}, more precisely different AVEVA PI System \cite{ref22} environments. Data from simulated instances were generated with OLGA \cite{ref23}, a dynamic multiphase flow simulator widely adopted by oil companies worldwide. Data from hand-drawn instances were created by Petrobras experts using a digital tool developed exclusively for this purpose.

Each type of instance, and therefore each data source, required the development of its own method for acquiring and labeling data. The common characteristics among the three developed methods are described in the following subsection. The particularities of each method are detailed in the subsequent subsections.

\clearpage

\subsection{Common Characteristics Among the Methods}

All instances, regardless of their type, are related to satellite-type offshore oil-producing wells that operate without manifold \cite{ref24}. These wells can alternate between different lifting methods \cite{ref25} over time, utilizing either the natural method or an artificial lifting method. The natural method is employed when the reservoir pressure is sufficient to produce hydrocarbons at a commercially viable rate without the need for additional energy. When reservoir pressure is insufficient, an artificial lifting method is required to introduce extra energy into the system and maintain production.

\begin{figure}[hbt!]
\centering
\includegraphics[width=1.0\textwidth]{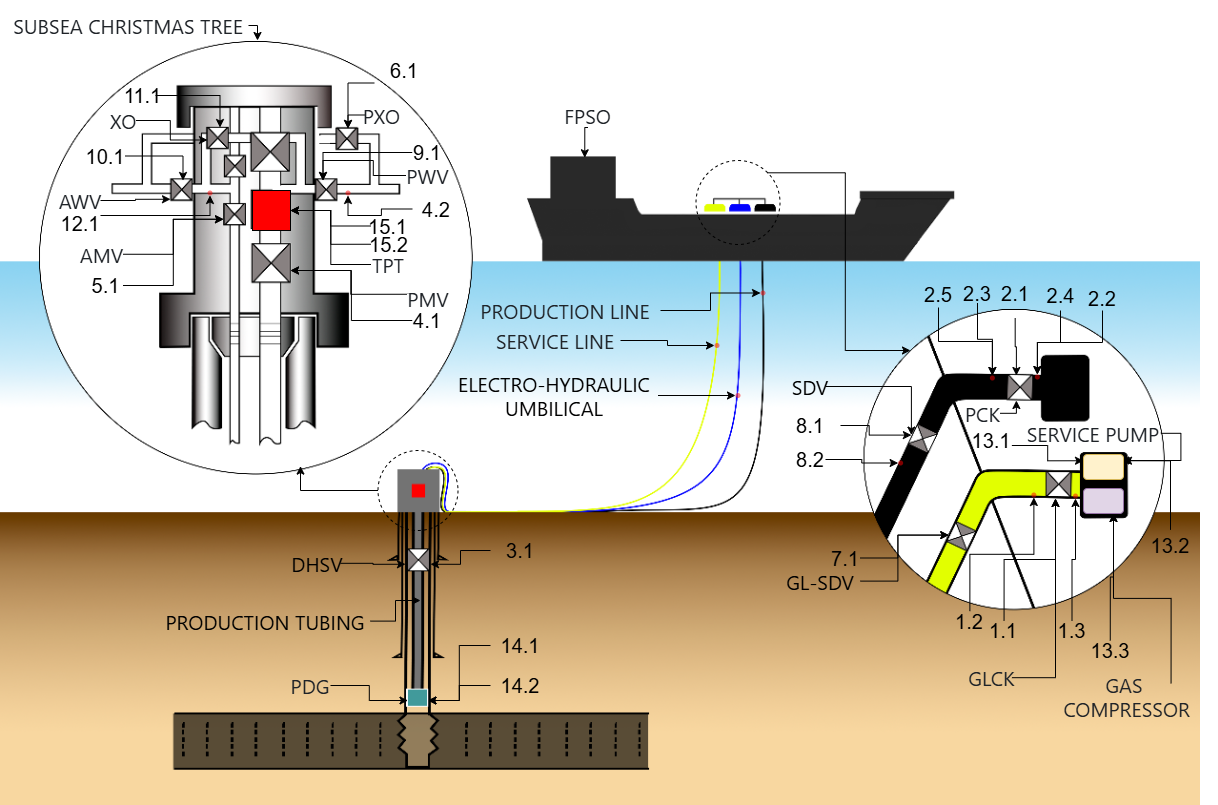}
\caption{Diagram representing the considered scenario when designing the 3W Dataset 2.0.0.}
\label{fig:diagram}
\end{figure}

Fig. \ref{fig:diagram} presents a diagram illustrating the scenario considered during the design of the 3W Dataset 2.0.0. This diagram covers only the components necessary for a good understanding of how this dataset was conceived. In summary, it depicts the production platform, the well, the subsea Christmas tree \cite{ref26}, the production and service lines, the electro-hydraulic umbilical, as well as sensors and valves.

The 3W Dataset 2.0.0 contains 27 variables present in all instances. Following the adopted definition of MTS, each variable is included in an instance even if no data has been obtained or generated for that variable in a specific instance. Table  \ref{tab:table2} provides the name of each variable, its representation, and its physical position within the scenario depicted in Fig. \ref{fig:diagram}.

\begin{table}[hbt!]
\centering
\caption{Details of the variables in the 3W Dataset 2.0.0.}
{\renewcommand{\arraystretch}{1.5}
\begin{tabular}{| l | l | c |}
\hline
\rowcolor[gray]{0.8} 
\multicolumn{1}{|c|}{\textbf{Name}} & 
\multicolumn{1}{c|}{\textbf{Description}} & 
\multicolumn{1}{c|}{\textbf{Position}} \\
\hline
ABER-CKGL & Opening of the GLCK (gas lift choke) & 1.1 \\
\hline
ABER-CKP & Opening of the PCK (production choke) & 2.1 \\
\hline
ESTADO-DHSV & State of the DHSV (downhole safety valve) & 3.1 \\
\hline
ESTADO-M1 & State of the PMV (production master valve) & 4.1 \\
\hline
ESTADO-M2 & State of the AMV (annulus master valve) & 5.1 \\
\hline
ESTADO-PXO & State of the PXO (pig-crossover) valve & 6.1 \\
\hline
ESTADO-SDV-GL & State of the gas lift SDV (shutdown valve) & 7.1 \\
\hline
ESTADO-SDV-P & State of the production SDV & 8.1 \\
\hline
ESTADO-W1 & State of the PWV (production wing valve) & 9.1 \\
\hline
ESTADO-W2 & State of the AWV (annulus wing valve) & 10.1 \\
\hline
ESTADO-XO & State of the XO (crossover) valve & 11.1 \\
\hline
P-ANULAR & Pressure in the well annulus & 12.1 \\
\hline
P-JUS-BS & Downstream pressure of the SP (service pump) & 13.1 \\
\hline
P-JUS-CKGL & Downstream pressure of the GLCK & 1.2 \\
\hline
P-JUS-CKP & Downstream pressure of the PCK & 2.2 \\
\hline
P-MON-CKGL & Upstream pressure of the GLCK & 1.3 \\
\hline
P-MON-CKP & Upstream pressure of the PCK & 2.3 \\
\hline
P-MON-SDV-P & Upstream pressure of the production SDV & 8.2 \\
\hline
P-PDG & Downhole pressure at the PDG (permanent downhole gauge) & 14.1 \\
\hline
PT-P & Subsea Xmas-tree pressure downstream of the PWV in the production line & 4.2 \\
\hline
P-TPT & Subsea Xmas-tree pressure at the TPT (temperature and pressure transducer) & 15.1 \\
\hline
QBS & Flow rate at the SP & 13.2 \\
\hline
QGL & Gas lift flow rate & 13.3 \\
\hline
T-JUS-CKP & Downstream temperature of the PCK & 2.4 \\
\hline
T-MON-CKP & Upstream temperature of the PCK & 2.5 \\
\hline
T-PDG & Downhole temperature at the PDG & 14.2 \\
\hline
T-TPT & Subsea Xmas-tree temperature at the TPT & 15.2 \\
\hline
\end{tabular}}
\label{tab:table2}
\end{table}

Note that some variable names contain terms or acronyms in Portuguese. For example: ABER = abertura = opening; CKGL = GLCK; CKP = PCK; ESTADO = state; ANULAR = annulus; JUS = jusante = downstream; and MON = montante = upstream. Translating all these names into English is a pending issue that will be resolved in future versions of 3W Dataset.

All instances were generated with observations recorded every 1 second, resulting in a fixed sampling frequency of 1 Hz.

All variables corresponding to a given physical quantity (type of variable) are expressed using the same measurement unit, as specified in Table  \ref{tab:table3}.

\begin{table}[hbt!]
\centering
\caption{Physical quantities and their measurement units.}
{\renewcommand{\arraystretch}{1.5}
\begin{tabular}{| l | l |}
\hline
\rowcolor[gray]{0.8} 
\multicolumn{1}{|c|}{\textbf{Physical Quantity}} & 
\multicolumn{1}{c|}{\textbf{Measurement Unit}} \\
\hline
Choke opening & \% \\
\hline
Flow rate & m$^3$/s \\
\hline
Pressure & Pa \\
\hline
Temperature & $^\circ$C \\
\hline
Valve state & Non-dimensional: 0 for closed, 1 for open, and 0.5 for any other state \\
\hline
\end{tabular}}
\label{tab:table3}
\end{table}

The labeling process applied to all instances of the 3W Dataset 2.0.0 resulted in two types of labels: class labels and state labels. Class labels are associated with normal conditions or the occurrence of undesirable events, while state labels are related to the operational status of the respective well.

The codes corresponding to the class labels are detailed in Table  \ref{tab:table4}. Any of these codes can be assigned to any observation from any instance. Codes 101 to 109 represent transient conditions between normal operation and steady states associated with undesirable events. It is important to note that not all undesirable events have corresponding transient conditions. When the well condition in terms of normality or the occurrence of an undesirable event is unknown at any time, the associated observation is labeled with the class label Unknown (code = NaN = Not a Number). Additionally, each instance, as a whole, is associated with a single steady state code (from 1 to 9) that corresponds to at least part of its observations. This code is referred to as the type of event.

\begin{table}[hbt!]
\centering
\caption{Class labels and their codes.}
{\renewcommand{\arraystretch}{1.5}
\begin{tabular}{| l | c | c |}
\hline
\rowcolor[gray]{0.8} 
\multicolumn{1}{|c|}{\textbf{Class Label}} & 
\multicolumn{1}{c|}{\textbf{Steady State Code}} & 
\multicolumn{1}{c|}{\textbf{Transient Condition Code}} \\
\hline
Normal Operation & 0 & - \\
\hline
Abrupt Increase of BSW & 1 & 101 \\
\hline
Spurious Closure of DHSV & 2 & 102 \\
\hline
Severe Slugging & 3 & - \\
\hline
Flow Instability & 4 & - \\
\hline
Rapid Productivity Loss & 5 & 105 \\
\hline
Quick Restriction in PCK & 6 & 106 \\
\hline
Scaling in PCK & 7 & 107 \\
\hline
Hydrate in Production Line & 8 & 108 \\
\hline
Hydrate in Service Line & 9 & 109 \\
\hline
Unknown & NaN & - \\
\hline
\end{tabular}}
\label{tab:table4}
\end{table}

The following are succinct descriptions of each type of event:

\begin{itemize}
    \item Abrupt Increase of BSW: a sudden rise in Basic Sediment and Water ratio, potentially leading to flow assurance issues, reduced oil production, and operational challenges;
    \item Spurious Closure of DHSV: an unexpected closure of Downhole Safety Valve, risking production losses;
    \item Severe Slugging: periodic and intense flow instability that can damage equipment and disrupt operations;
    \item Flow Instability: non-periodic flow disturbances that, if unaddressed, may escalate into severe slugging;
    \item Rapid Productivity Loss: a sudden decrease in well productivity, often driven by changes in system properties;
    \item Quick Restriction in PCK: a rapid and significant restriction in the production choke valve, typically caused by operational challenges;
    \item Scaling in PCK: formation of inorganic deposits in the production choke valve, reducing oil and gas production;
    \item Hydrate in Production Line: formation of crystalline compounds (hydrates) that can block pipelines, leading to significant production losses and high unblocking costs;
    \item Hydrate in Service Line: similar to Hydrate in Production Line, but occurring in the service line, leading to distinct signature patterns in operational data.
\end{itemize}

For further details on these events, please refer to Vargas et al. \cite{ref09}.

The codes associated with the state labels are detailed in Table  \ref{tab:table5}. These codes can be associated with any observation from any instance, but they are not associated with any instance as a whole. If the well's operating condition is unknown at any given time, the respective observation is assigned the state label Unknown (code = NaN).

\begin{table}[hbt!]
\centering
\caption{State labels and their codes.}
{\renewcommand{\arraystretch}{1.5}
\begin{tabular}{| l | c |}
\hline
\rowcolor[gray]{0.8} 
\multicolumn{1}{|c|}{\textbf{State Label}} & 
\multicolumn{1}{c|}{\textbf{Transient Condition Code}} \\
\hline
Open & 0 \\
\hline
Shut-In & 1 \\
\hline
Flushing Diesel & 2 \\
\hline
Flushing Gas & 3 \\
\hline
Bullheading & 4 \\
\hline
Closed With Diesel & 5 \\
\hline
Closed With Gas & 6 \\
\hline
Restart & 7 \\
\hline
Depressurization & 8 \\
\hline
Unknown & NaN \\
\hline
\end{tabular}}
\label{tab:table5}
\end{table}

Each operational status is defined based on the configuration of the valves and the nature of recent and/or ongoing operations. In summary, the statuses are as follows:

\begin{itemize}
    \item Open: all production valves (M1, W1, SDV-P, and PCK) are open, and auxiliary valves (PXO, and XO) are closed. The well is producing under regular condition;
    \item Shut-in: at least one valve in the production path is closed. The well is closed, so it is not producing;
    \item Flushing Diesel: at least one wellhead valve is closed, PXO or XO is open, and diesel is injected. The well is closed, and a diesel circulation operation from service line to production line is being executed;
    \item Flushing Gas: at least one wellhead valve is closed, PXO or XO is open, and gas is injected. The well is closed, and a gas circulation operation from service line to production line is being executed;
    \item Bullheading: all production valves are open, and diesel or gas is injected from the topside through the production line. The well is closed, and the production line is being pressurized from topside by diesel or gas to push down all the production fluids back into the well;
    \item Closed With Diesel: at least one production valve is closed, and the previous state was Flushing Diesel or Bullheading. The well is closed, and the majority of the production line is filled with diesel. This condition mitigates hydrates risk from occurrence;
    \item Closed With Gas: at least one production valve is closed, and the previous state was Flushing Gas. The well is closed, and the majority of the production line is filled with natural gas. This condition mitigates hydrates risk from occurrence;
    \item Restart: following a shut-in, all production valves are reopened. The well is recently opened, so it is in a transient period before it gets to the Open operational status;
    \item Depressurization: following a shut-in, SDV-P and PCK are open, while M1, W1, PXO, and XO remain closed. Production line is depressurized in order to mitigate hydrate risk from occurrence.
\end{itemize}

Understanding and correctly identifying these operational statuses is critical in monitoring and preventing undesirable events. Some important additional explanations are as follows:

\begin{itemize}
    \item Open state denotes continuous production with all key production path valves open and no auxiliary operations ongoing;
    \item All operational statuses other than Open are associated with well Shut-in procedures;
    \item Closed With Diesel state occurs following successful Flushing Diesel or Bullheading, during which the well is filled with diesel (a condition in which hydrate formation is highly unlikely);
    \item Bullheading is performed by injecting fluid (diesel or gas) directly from the topside into the production line under pressure, pushing the wellbore fluids back into the reservoir. This operation is used to ensure that a non-hydrate-forming fluid fills the system, often as a preventive action against hydrate plugs;
    \item Flushing Diesel, in contrast, is performed by circulating diesel through the service line and into the production flowline, displacing the original production fluids along the flowline path, typically toward the topside facilities. This operation ensures that hydrate-prone fluids are replaced with a more stable medium, reducing the chances of hydrate formation during shut-in;
    \item Closed With Gas state occurs following a Flushing Gas operation, which is performed by injecting and circulating gas (usually dry gas) via the service line into the production line. This operation also serves to remove liquid hydrocarbons or water that could contribute to hydrate formation, creating a drier environment inside the pipeline;
    \item Additionally, historical operational context can be valuable; for instance, depressurizing the production system is a typical measure performed shortly after the well is shut-in, in order to seek for thermodynamic conditions less favorable to the formation of hydrates;
    \item Depressurization state may indicate that subcooling levels were reduced, potentially mitigating hydrate risks during extended shut-ins.
\end{itemize}

\subsection{Method Relating to Real Instances}

The particularities of the method developed for real data are listed below.

\begin{figure}[hbt!]
\centering
\includegraphics[width=0.67\textwidth]{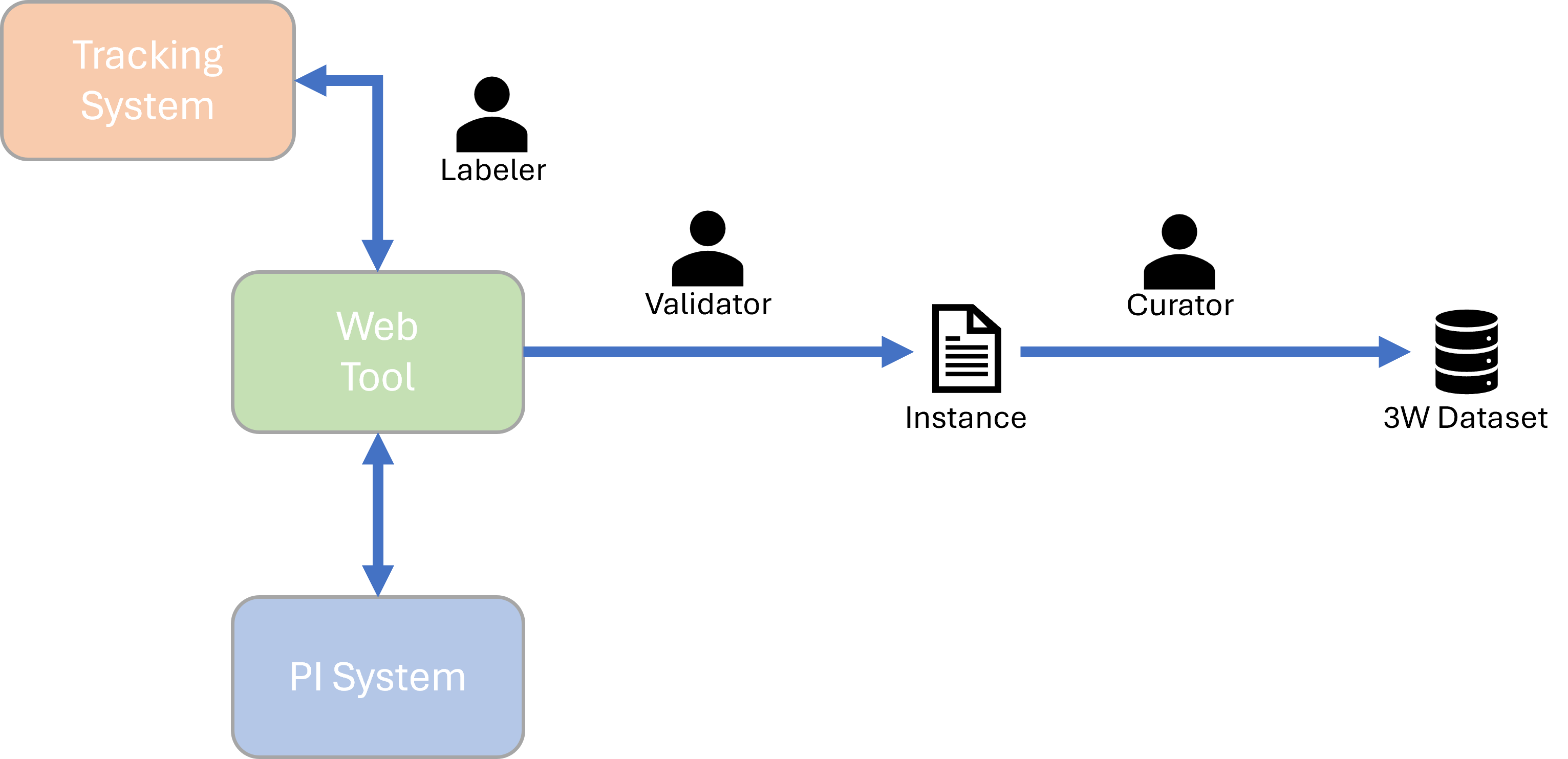}
\caption{Illustration of the labeling process of real instances.}
\label{fig:labeling_process}
\end{figure}

\begin{figure}[hbt!]
\centering
\includegraphics[width=1.0\textwidth]{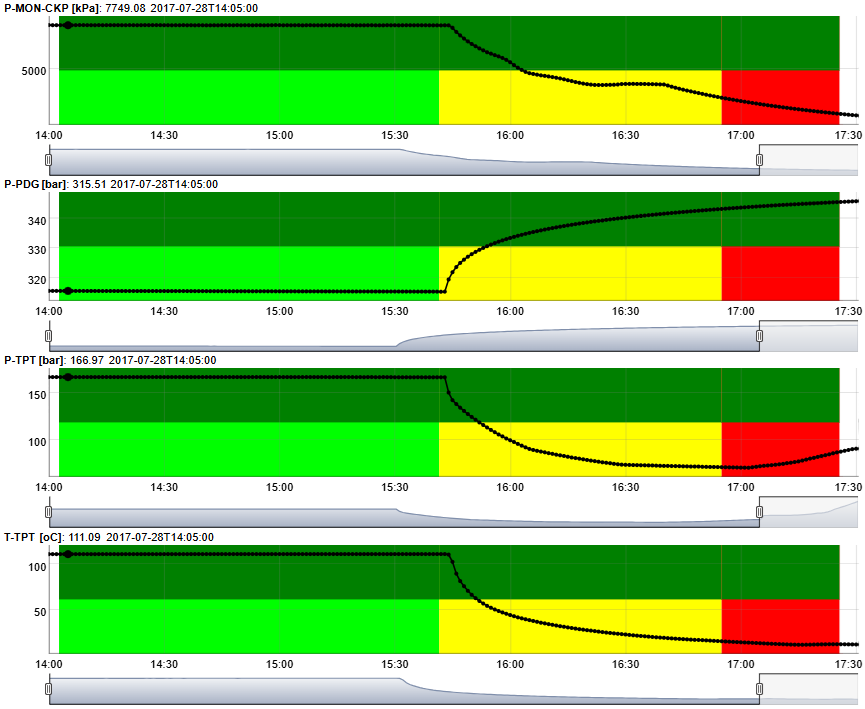}
\caption{Example of a real Spurious Closure of DHSV instance in which pressure increase is observed in sensors located upstream of the DHSV (PDG) and pressure decreases are observed in sensors located downstream of the DHSV (TPT and MON-CKP). This example has 3 periods which were labeled with the Petrobras' Web Tool as follows: light green (1st period = Normal Operation): class label = 0; yellow (2nd period = Transient Condition): class label = 102; red (3rd period = Steady State): class label = 2; dark green (Open): state label = 0.}
\label{fig:spurious_closure_dhsv}
\end{figure}

\begin{figure}[hbt!]
\centering
\includegraphics[width=1.0\textwidth]{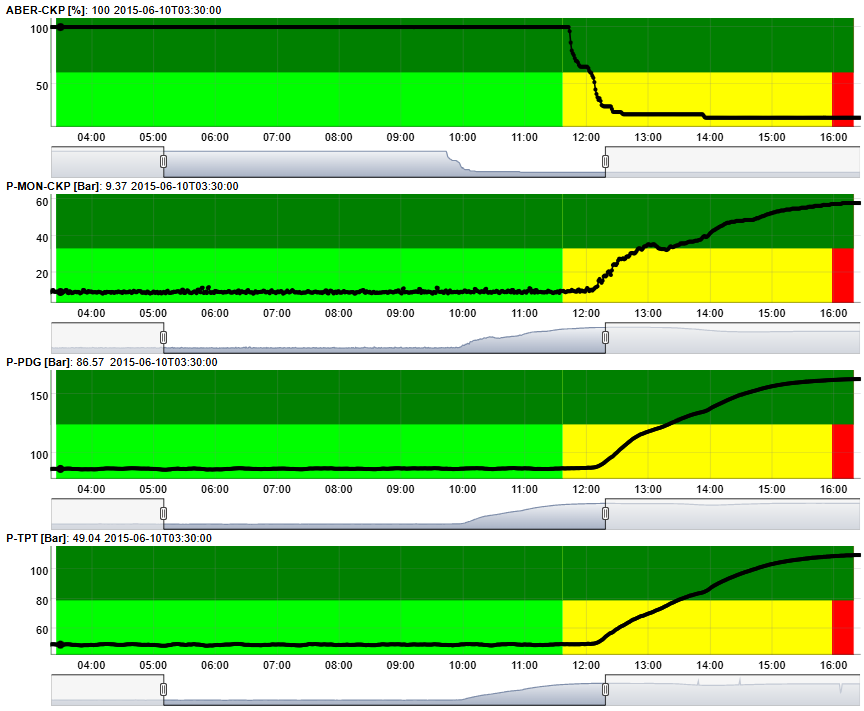}
\caption{Example of a real Quick Restriction in PCK instance in which pressure increases are observed in sensors located in TPT, PDG, and MON-CKP. This example has 3 periods which were labeled with the Petrobras' Web Tool as follows: light green (1st period = Normal Operation): class label = 0; yellow (2nd period = Transient Condition): class label = 106; red (3rd period = Steady State): class label = 6; dark green (Open): state label = 0.}
\label{fig:quick_restriction_pck}
\end{figure}

\begin{figure}[hbt!]
\centering
\includegraphics[width=1.0\textwidth]{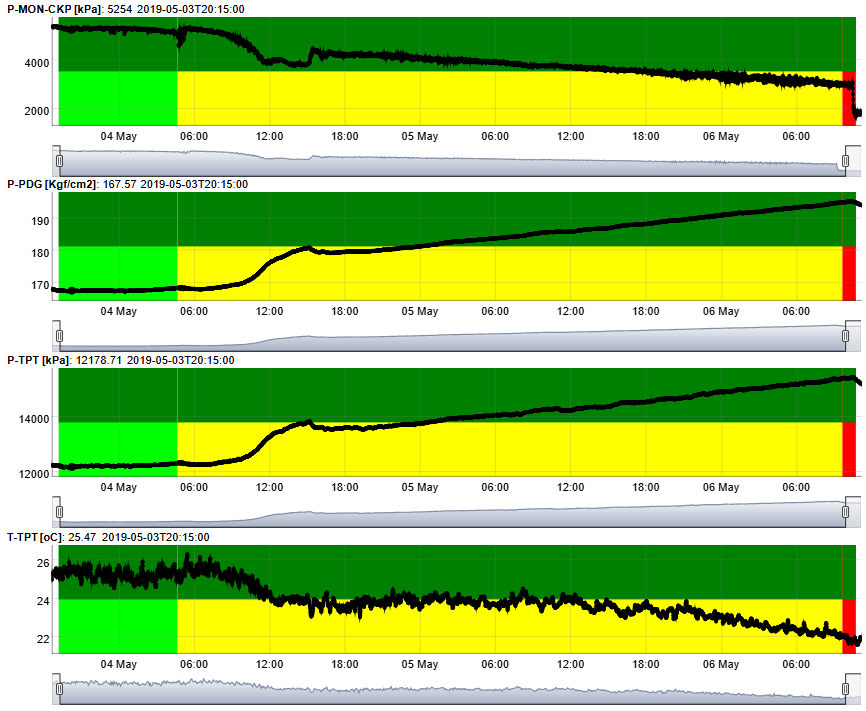}
\caption{Example of a real Hydrate in Production Line instance in which pressure increases are observed in sensors located upstream of the production line (PDG and TPT) and pressure decrease is observed in sensor located downstream of the production line (MON-CKP). This example has 3 periods which were labeled with the Petrobras' Web Tool as follows: light green (1st period = Normal Operation): class label = 0; yellow (2nd period = Transient Condition): class label = 108; red (3rd period = Steady State): class label = 8; dark green (Open): state label = 0.}
\label{fig:hydrate_production_line}
\end{figure}

\begin{figure}[hbt!]
\centering
\includegraphics[width=1.0\textwidth]{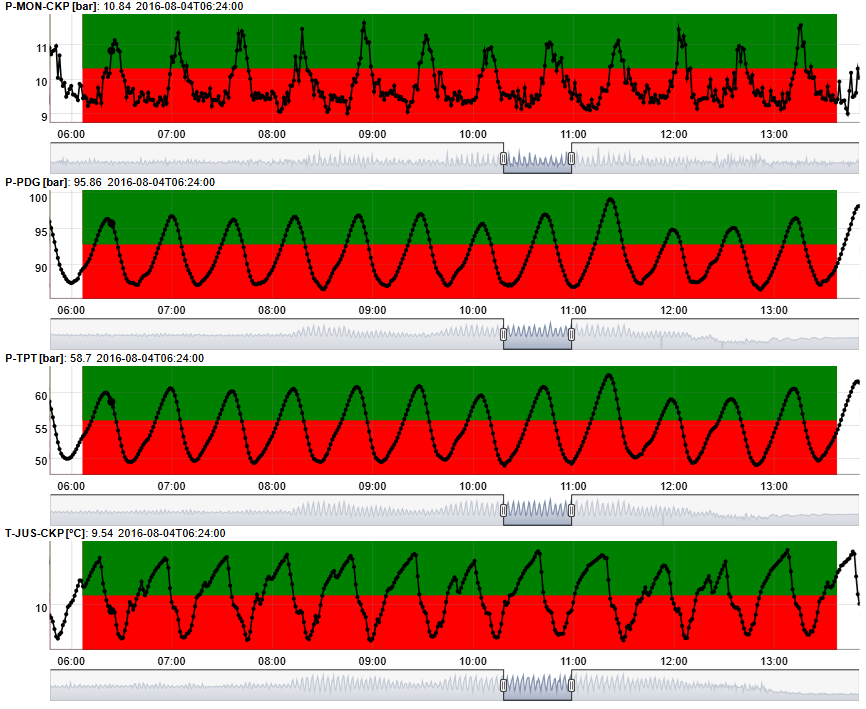}
\caption{Example of a real Severe Slugging instance in which pressure oscillations with amplitudes above 10 bar are observed in sensors located in PDG and TPT. This example has 1 period which was labeled with the Petrobras' Web Tool as follows: red (1st period = Steady State): class label = 3; dark green (Open): state label = 0.}
\label{fig:severe_slugging}
\end{figure}

\begin{figure}[hbt!]
\centering
\includegraphics[width=1.0\textwidth]{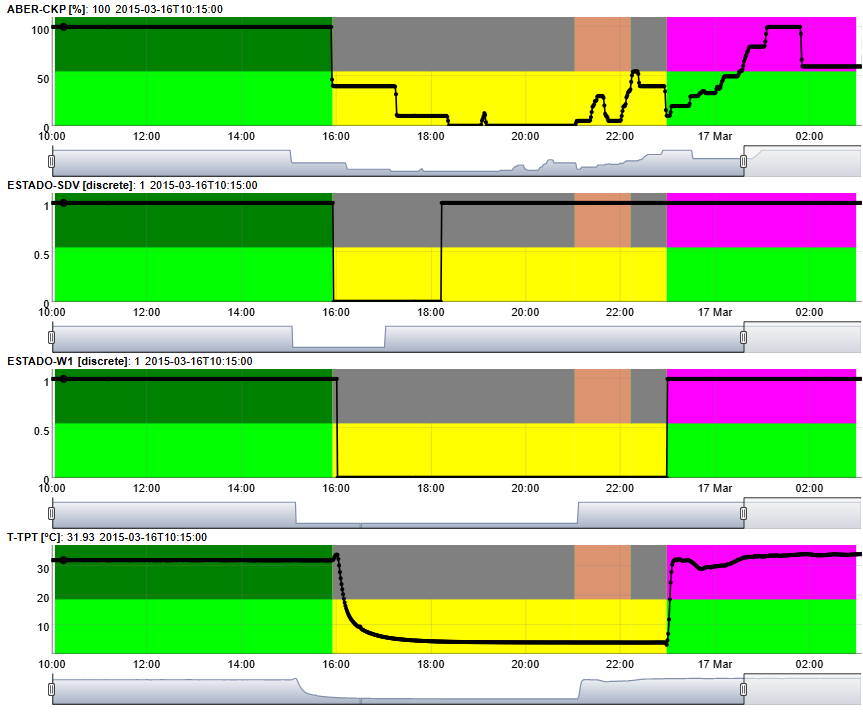}
\caption{Example of a real Normal Operation instance in which depressurization has been carried out during a well shutdown, as shown by the opening of the PCK and SDV-P. This example has 1 period which was labeled with the Petrobras' Web Tool as follows: light green (1st period = Normal Operation): class label = 0; dark green (Open): state label = 0; gray (Shut-In): state label = 1; salmon (Depressurization): state label = 8; magenta (Restart): state label = 7.}
\label{fig:normal_operation}
\end{figure}

\begin{itemize}
    \item Data acquisition:
    \begin{enumerate}
        \item Data is sourced from Petrobras' PIMS \cite{ref21}, specifically from different AVEVA PI System \cite{ref22} environments;
        \item Linear interpolation provided by the AVEVA PI System is applied to simulate a fixed acquisition frequency of 1 Hz;
        \item No preprocessing is performed for frozen variables or missing values;
        \item Data is converted to standard measurement units.
    \end{enumerate}
    \item Data labeling:
    \begin{enumerate}
        \item Historical records are mapped in a Tracking System to ensure accessibility and traceability for labelers (experts from Petrobras and its partners);
        \item Labelers use a Petrobras' Web Tool developed to label and export real data to the 3W Dataset;
        \item An expert committee (Validators) reviews the labeling, suggests potential adjustments, and validates the assigned labels;
        \item A Curator (a 3W Dataset specialist) finalizes the labeling process by updating the tool to mark validated events (instances) as labeled, enabling their exportation and inclusion in the 3W Dataset;
        \item Fig. \ref{fig:labeling_process} illustrates this labeling process; 
        \item Figs. \ref{fig:spurious_closure_dhsv} to \ref{fig:normal_operation} provide examples of labeled real instances. In these examples, only part of the variables are presented (before any standardization of measurement unit).
    \end{enumerate}
\end{itemize}

The main limitations of this method are:

\begin{enumerate}
    \item It covers only events that occurred in real life and had archived records;
    \item Contextualization (mappings between variables and tags in PIMS) was not verified;
    \item Original measurement units configured in PIMS (before conversions) were not verified.
\end{enumerate}

\clearpage

\clearpage

\clearpage

\clearpage

\clearpage

\clearpage

\subsection{Method Relating to Simulated Instances}

The particularities of the method developed for simulated data are listed below.

\begin{itemize}
    \item Data generation:
    \begin{enumerate}
        \item A matrix was designed for simulations with OLGA \cite{ref23}, gradually varying the main parameters of the considered scenario;
        \item Simulations were run, and results that did not converge (e.g., wells that did not produce normally) were filtered out;
        \item Time series were automatically extracted from TPL files generated by OLGA, corresponding to successful simulations;
        \item Time series are perfectly periodic;
        \item No frozen variables or missing values are present;
        \item Variables are represented with standardized measurement units;
        \item No noise is introduced in the time series.
    \end{enumerate}        
    \item Data labeling:
    \begin{enumerate}
        \item Labeling is fully automated based on simulation results, including transient condition periods.
    \end{enumerate}
\end{itemize}

The main limitations of this method are:

\begin{enumerate}
    \item The method uses a single phenomenological model in OLGA associated with a single well;
    \item A simplified simulation strategy was employed. For example: hydrate formation was simulated using a valve with a linear closing percentage;
    \item Some variables were not included in the simulations, resulting in all their values being considered missing across all simulated instances.
\end{enumerate}

\subsection{Method Relating to Hand-Drawn Instances}

The particularities of the method developed for hand-drawn data are listed below.

\begin{figure}[hbt!]
\centering
\includegraphics[width=1.0\textwidth]{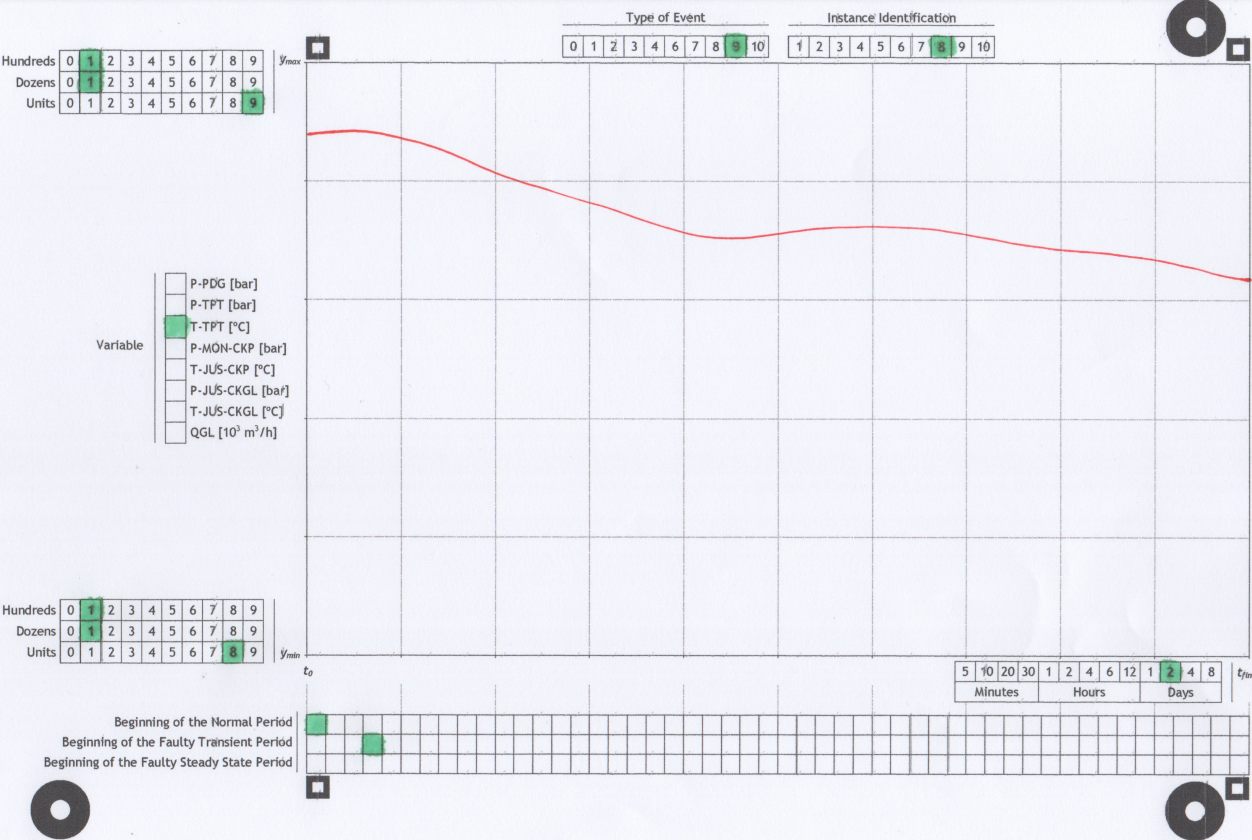}
\caption{Example of a variable drawn and labeled by hand by an expert from Petrobras in the tool developed exclusively for generating instances for the 3W Dataset.}
\label{fig:variable_drawn_and_labeled_by_hand}
\end{figure}

\begin{itemize}
    \item Data generation:
    \begin{enumerate}
        \item A proprietary tool based on image processing was developed exclusively for generating hand-drawn data for the 3W Dataset;
        \item Each variable was hand-drawn on its own chart by an expert from Petrobras. An example is shown in Fig. \ref{fig:variable_drawn_and_labeled_by_hand};
        \item Time series were automatically digitized by scanning paper-printed graphs;
        \item Time series are perfectly periodic;
        \item No frozen variables or missing values are present;
        \item Variables are represented with standardized measurement units.
    \end{enumerate}  
    \item Data labeling:
    \begin{enumerate}
        \item Labels were derived from expert markings on paper-printed graphs, including transient condition periods.
    \end{enumerate}
\end{itemize}

The main limitations of this method are:

\begin{enumerate}
    \item The method covers only well-known signatures of the considered undesirable events;
    \item There is potential for expert bias when drawing and labeling time series;
    \item The process depends on the manual dexterity of the experts.
\end{enumerate}

\section{Data Records}

The 3W Dataset 2.0.0 is licensed under CC BY 4.0 \cite{ref27} and is publicly available at the following Figshare \cite{ref28} address: \href{https://doi.org/10.6084/m9.figshare.29205836.v1}{https://doi.org/10.6084/m9.figshare.29205836.v1}.

An overview of this dataset is presented in the next subsection, which covers the main quantities and statistics. The subsequent subsection details how the data records are structured.

\subsection{3W Dataset 2.0.0 Overview}

The quantities of instances that compose the 3W Dataset 2.0.0, categorized by type of instance and type of event, are presented in Table  \ref{tab:table6}.

\begin{table}[hbt!]
\centering
\caption{Quantities of instances that compose the 3W Dataset 2.0.0. Values in brackets represent the quantities from version 1.0.0 that differ in the current version.}
{\renewcommand{\arraystretch}{1.5}
\begin{tabular}{| l | c | c | c | c |}
\hline
\rowcolor[gray]{0.8} 
\multicolumn{1}{|c|}{\textbf{Type of Event}} & 
\multicolumn{1}{c|}{\textbf{Real}} & 
\multicolumn{1}{c|}{\textbf{Simulated}} & 
\multicolumn{1}{c|}{\textbf{Hand-Drawn}} & 
\multicolumn{1}{c|}{\textbf{Total}} \\
\hline
0 - Normal Operation & 594 (597) & 0 & 0 & \textbf{594 (597)} \\
\hline
1 - Abrupt Increase of BSW & 4 (5) & 114 & 10 & \textbf{128 (129)} \\
\hline
2 - Spurious Closure of DHSV & 22 & 16 & 0 & \textbf{38} \\
\hline
3 - Severe Slugging & 32 & 74 & 0 & \textbf{106} \\
\hline
4 - Flow Instability & 343 (344) & 0 & 0 & \textbf{343 (344)} \\
\hline
5 - Rapid Productivity Loss & 11 (12) & 439 & 0 & \textbf{450 (451)} \\
\hline
6 - Quick Restriction in PCK & 6 & 215 & 0 & \textbf{221} \\
\hline
7 - Scaling in PCK & 36 (4) & 0 & 10 & \textbf{46 (14)} \\
\hline
8 - Hydrate in Production Line & 14 (3) & 81 & 0 & \textbf{95 (84)} \\
\hline
9 - Hydrate in Service Line & 57 (0) & 150 (0) & 0 & \textbf{207 (0)} \\
\hline
\textbf{Total} & \textbf{1119 (1025)} & \textbf{1089 (939)} & \textbf{20} & \textbf{2228 (1984)} \\
\hline
\end{tabular}}
\label{tab:table6}
\end{table}

\begin{figure}[hbt!]
\centering
\includegraphics[width=1.0\textwidth]{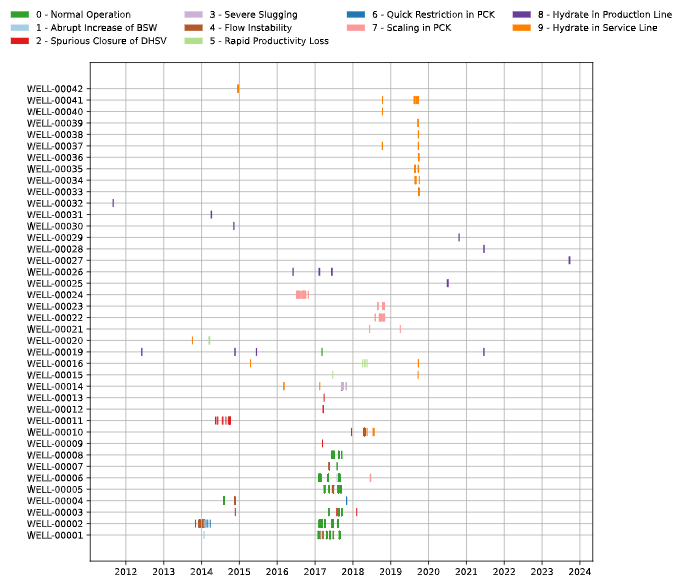}
\caption{Scatter map with all the real instances in the 3W Dataset 2.0.0.}
\label{fig:scatter_map}
\end{figure}

Fig. \ref{fig:scatter_map} shows a scatter map containing all the real instances. The oldest instance occurred in mid-2011, while the most recent occurred in mid-2023. In addition to the total number of wells considered (42), this map provides an overview of the distribution of instances over time and across wells.

Scatter map with all the real instances in the 3W Dataset 2.0.0.

The 3W Dataset's main statistics related to inherent difficulties of real data are presented in Table  \ref{tab:table7}.

\begin{table}[hbt!]
\centering
\caption{The 3W Dataset's main statistics related to inherent difficulties of real data.}
{\renewcommand{\arraystretch}{1.5}
\begin{tabular}{| l | c | c |}
\hline
\rowcolor[gray]{0.8}
\multicolumn{1}{|c|}{\textbf{Statistic}} & 
\multicolumn{1}{c|}{\textbf{Amount}} & 
\multicolumn{1}{c|}{\textbf{Percentage}} \\
\hline
Missing Variables & 41,109 & 65.90\% of 62,384 \\
\hline
Frozen Variables & 6,095 & 9.77\% of 62,384 \\
\hline
Unlabeled Observations & 4,028,400 & 5.26\% of 76,587,318 \\
\hline
\end{tabular}}
\label{tab:table7}
\end{table}

\subsection{3W Dataset 2.0.0 Structure}

In the root of the directory containing the dataset, there is a file called dataset.ini, which specifies properties of the 3W Dataset 2.0.0. The proposal is that all users concentrate their searches for these properties in this file.

The data itself is organized into subdirectories dedicated to each type of event. The name of each directory corresponds to the code associated with each type of event (see Table  \ref{tab:table6}).

Each instance is stored in its own Apache Parquet file \cite{ref29}, or simply Parquet file. Parquet is an open source, column-oriented data file format designed for efficient data storage and retrieval. It supports high-performance compression and encoding schemes, allowing it to handle complex data in bulk. It is also supported in many programming languages and analytics tools.

The logic used to formulate file names depends on the type of instance.

\begin{itemize}
    \item Real Instances:  
    \begin{itemize}
        \item File names follow the format WELL-[incremental ID]\_[timestamp of oldest observation].parquet. Example: WELL-00014\_20170917140000.parquet;
        \item Each real well is associated with a unique ID, regardless of the type of event (subdirectory). As each real well can give rise to one or multiple instances, the [timestamp of oldest observation] ensures unique identification.
    \end{itemize}          
    \item Simulated Instances:
    \begin{itemize}    
        \item File names follow the format SIMULATED\_[incremental ID].parquet. Example: SIMULATED\_00072.parquet;
        \item The incremental ID starts at 1 for each type of event (subdirectory) and is sufficient to uniquely identify all simulated instances.
    \end{itemize}          
    \item Hand-Drawn Instances:
    \begin{itemize}    
        \item File names follow the format DRAWN\_[incremental ID].parquet. Example: DRAWN\_00007.parquet;
        \item The incremental ID starts at 1 for each type of event (subdirectory) and is sufficient to uniquely identify all hand-drawn instances.
    \end{itemize}
\end{itemize}

All Parquet files are created with the Pyarrow engine \cite{ref30} and the Brotli compression \cite{ref31}. These choices provide a good compromise between compression ratio and read performance.

The timestamp vector of each instance is used as the index in the corresponding Parquet file. All timestamps are represented in the format 'YYYY-MM-DD HH:MM:SS'.

All variables and labels are stored as columns in Parquet files, variables as float and labels as Int64 (not int64).

\section{Technical Validation}

Several carefully incorporated features in the methods described in the Methods Section ensure the high-technical quality of the 3W Dataset 2.0.0. The most relevant features are the following:

\begin{itemize}
    \item Real instances: preservation of real data characteristics, labeling by experts, and validation by expert committee;
    \item Simulated instances: simulation models calibrated by experts, and systematized labeling;
    \item Hand-drawn instances: hand-drawn graphs by experts, and systematized labeling.
\end{itemize}

The experts mentioned here are all from Petrobras and its partners.

\begin{figure}[hbt!]
\centering
\includegraphics[width=0.9\textwidth]{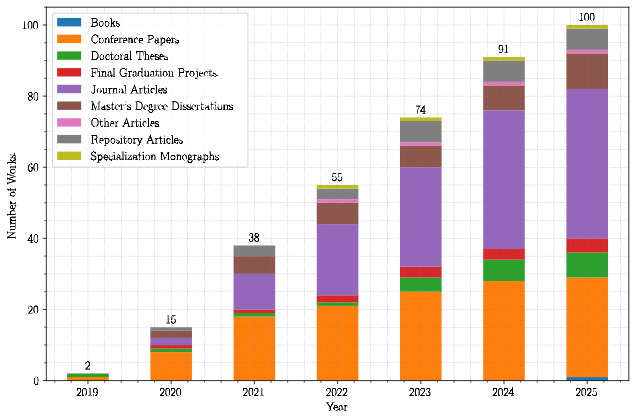}
\caption{The quantitative annual progress of different types of publications that cite the 3W Dataset.}
\label{fig:citation_progress}
\end{figure}

The quantity and the diversity of the works developed and published by the 3W Community demonstrate the relevance and the high-technical quality of the 3W Dataset. This framework is composed of books, conference papers, doctoral theses, final graduation projects, journal articles, master's degree dissertations, repository articles, and specialization monographs. The quantitative annual progress of these publications to date is shown in Fig. \ref{fig:citation_progress}. Over the years, the adoption of the 3W Dataset by the 3W Community has steadily increased, highlighting its relevance, accessibility, and practical applicability in advancing AEM systems. More information on these works can be found in the 3W Project repository \cite{ref17}.

The quantitative annual progress of different types of publications that cite the 3W Dataset.

\section{Usage Notes}

This data article describes the 3W Dataset 2.0.0, the current publicly available version, and summarizes its evolution from version 1.0.0, as previously detailed by Vargas et al. \cite{ref09}.

The 3W Dataset 2.0.0 was generated with Python 3.10 \cite{ref20} code, using resources mainly provided by the Pandas 1.5 \cite{ref32}, and Pyarrow 19.0 \cite{ref30} packages. 

Designed for cross-platform compatibility, the 3W Dataset 2.0.0 can be explored using virtually any programming language. However, regardless of the language, the Apache Parquet files \cite{ref29} must be accessed using the Pyarrow engine \cite{ref30} alongside Brotli compression \cite{ref31}.

When comparing results obtained in different works, it is important to distinguish which versions of the 3W Dataset were used. Due to substantial differences among the published versions, certain comparisons may pose challenges and require careful consideration.

The main updates introduced in version 2.0.0 compared to version 1.0.0 are summarized below. More details can be found in the release notes available in the 3W Project repository \cite{ref17}.

\begin{enumerate}
    \item The dataset structure has been significantly revised. Data is now stored in Parquet files instead of CSV files;
    \item One variable (T-JUS-CKGL) has been discontinued, and 20 others have been added, resulting in 27 variables in the current version;
    \item One new type of undesirable event has been added: Hydrate in Service Line;
    \item The number of real instances has increased by 94;
    \item The number of simulated instances has increased by 150;
    \item The number of real wells covered has doubled, increasing from 21 to 42;
    \item A new label, the state label, has been added;
    \item More labeled observations have been incorporated across several instances;
    \item No significant changes have been made to simulated or hand-drawn instances. All 20 new variables were incorporated into these instances with missing values.
\end{enumerate}

The operational status of a well, represented by the state label, is strongly correlated with the values of its variables. This allows algorithms to be applied to the 3W Dataset for quantifying these relationships. Additionally, the state label can be used to select data for specific training purposes.

\section{Code Availability}

The 3W Project also develops and maintains the 3W Toolkit, a publicly available software package, hosted in its Git repository \cite{ref17}. Written in Python 3 \cite{ref20} and licensed under Apache 2.0 \cite{ref33}, this toolkit is designed to facilitate and encourage exploration of the 3W Dataset as well as the development and evaluation of comparative approaches. It provides a range of features for loading data, visualizing time series, calculating metrics, and more. Python practitioners can therefore use and adapt the 3W Toolkit to speed up their analysis with the 3W Dataset 2.0.0.

An in-depth description of the 3W Toolkit is beyond the scope of this data article, but its documentation and examples of use can be found in the 3W Project repository \cite{ref17}.

\section{Acknowledgements}

The authors would like to thank Petróleo Brasileiro S.A. (Petrobras) for providing all the necessary resources for the preparation and publication of this data article.

\section{Author Contributions}

R.V. lead this project, labeled data, prepared this manuscript drafts version, and incorporated contributions from all co-authors to produce the submitted and published versions. A.J. lead this project, labeled data, added content, and reviewed texts. C.L. reviewed texts. C.M. added content, and reviewed texts. E.J. provided references, and reviewed texts. F.V. reviewed texts. G.P. added content, created figures, provided references, and reviewed texts. I.O. provided references, and reviewed texts. J.A. labeled data, merged contributions, and reviewed texts. J.B. added content, created figures, provided references, and reviewed texts. J.C. added content, created figures, provided references, and reviewed texts. L.G. provided references, and reviewed texts. L.L. provided references, and reviewed texts. M.F. provided references, and reviewed texts. P.C. added content, and reviewed texts. F.B. labeled data, added content, created figures, provided references, and reviewed texts. J.C. labeled data, added content, created figures, provided references, and reviewed texts. M.S. labeled data, added content, created figures, provided references, and reviewed texts. R.B. labeled data, added content, created figures, provided references, and reviewed texts. R.P. labeled data, added content, and reviewed texts.

\end{document}